\documentclass[utf8]{arxiv}   %

\usepackage{url,hyperref,microtype,subcaption}
\usepackage[onehalfspacing]{setspace}

\newcommand{\tabitem}{~~\llap{\textbullet}~~}
\usepackage{booktabs}
\usepackage{color}

\def\keyFont{\fontsize{8}{11}\helveticabold }
\def\firstAuthorLast{Kafle {et~al.}} 
\def\Authors{Kushal Kafle\,$^{1,*}$, Robik Shrestha\,$^{1}$ and Christopher Kanan\,$^{1,2,3,*}$}
\def\Address{$^{1}$Center for Imaging Science, Rochester Institute of Technology, Rochester, USA \\
$^{2}$PAIGE, New York, USA \\
$^{3}$Cornell Tech, New York, USA}
\def\corrAuthor{Christopher Kanan}

\def\corrEmail{kanan@rit.edu}

\begin{document}
\onecolumn
\firstpage{1}

\title[Challenges and Prospects in Vision and Language Research]{Challenges and Prospects in Vision and Language Research} 

\author[\firstAuthorLast ]{\Authors} 
\address{} 
\correspondance{} 

\extraAuth{Kushal Kafle \\ kk6055@rit.edu}

\maketitle

\begin{abstract}
Language grounded image understanding tasks have often been proposed as a method for evaluating progress in artificial intelligence. Ideally, these tasks should test a plethora of capabilities that integrate computer vision, reasoning, and natural language understanding. However, rather than behaving as visual Turing tests, recent studies have demonstrated state-of-the-art systems are achieving good performance through flaws in datasets and evaluation procedures. We review the current state of affairs and outline a path forward.

\section{}


\tiny
 \keyFont{ \section{Keywords:} computer vision, natural language understanding, visual question answering, captioning, dataset bias, visual Turing test} 
\end{abstract}

\section{Introduction}
\label{sec:intro}


Advancements in deep learning and the availability of large-scale datasets have resulted in great progress in computer vision and natural language processing (NLP). Deep convolutional neural networks (CNNs) have enabled unprecedented improvements in classical computer vision tasks, e.g., image classification and object detection. Progress in many NLP tasks has been similarly swift. Building upon these advances, there is a push to attack new problems that enable concept comprehension and reasoning capabilities to be studied at the intersection of vision and language (V\&L) understanding. There are numerous applications for V\&L systems, including enabling the visually impaired to interact with visual content using language, human-computer interaction, and visual search. Human-robot collaboration would be greatly enhanced by giving robots understanding of human language to better understand the visual world.

However, the primary objective of many scientists working on V\&L problems is to have them serve as stepping stones toward a visual Turing test~\citep{geman2015visual}, a benchmark for progress in artificial intelligence (AI). Grounding visual processing using language can  provide a test-bed for goal-directed visual understanding, with language queries determining the task to be performed. V\&L tasks can demand many disparate computer vision and NLP skills to be used simultaneously. The same system may be required to simultaneously engage in reasoning, object recognition, attribute detection, and more. Most V\&L benchmarks capture only a fraction of the requirements of a rigorous visual Turing test; however, we argue that across V\&L tasks a rigorous evaluation should assess numerous scene understanding capabilities individually and give confidence that an algorithm is right for the right reasons. If it is possible to do well on a benchmark by only answering common easy questions, not looking at the image, or by merely guessing using spurious correlations, then it will not satisfy these requisites for a good test.

Many V\&L tasks have been proposed, including image and video captioning~\citep{Mao2014DeepCW,yu2016video}, visual question answering (VQA)~\citep{antol2015vqa,kafle2016review,Zhang_2016_CVPR,agrawal2017cvqa,agrawal2018don,kafle2017analysis}, referring expression recognition (RER)~\citep{kazemzadeh2014referitgame}, image retrieval~\citep{mezaris2003ontology,johnson2015image}, activity recognition~\citep{yatskar2016situation,Zhao2017MenAL}, and language-guided image generation~\citep{reed2016generative,han2017stackgan}. A wide variety of algorithms have been proposed for each of these tasks, producing increasingly better results across datasets. However, several studies have called into question the \textit{true} capability of these systems and the efficacy of current assessment methods~\citep{kafle2017analysis,cirik2018visual,madhyastha2018end}. Systems are heavily influenced by dataset bias and lack robustness to uncommon visual configurations~\citep{agrawal2017cvqa,kafle2017analysis,madhyastha2018end}, but these are often not measured and call into question the value of these benchmarks. These issues also impact system assessment and deployment. Systems can amplify spurious correlations between gender and potentially unrelated variables in V\&L problems~\citep{Hendricks2018WomenAS,Zhao2017MenAL}, resulting in the possibility of severe negative real-world impact.

In this article, we outline the current state of V\&L research. We identify the challenges of developing good algorithms, datasets, and evaluation metrics. We discuss issues unique to individual tasks as well as identify common shortcomings shared across V\&L benchmarks. We provide our perspective on  potential future directions for V\&L research, especially on the requisites for benchmarks to better serve as a visual Turing tests.

\section{A Brief Survey of V\&L Research}\label{sec:survey}

\begin{figure}[t]
\centering
\includegraphics[width=\textwidth]{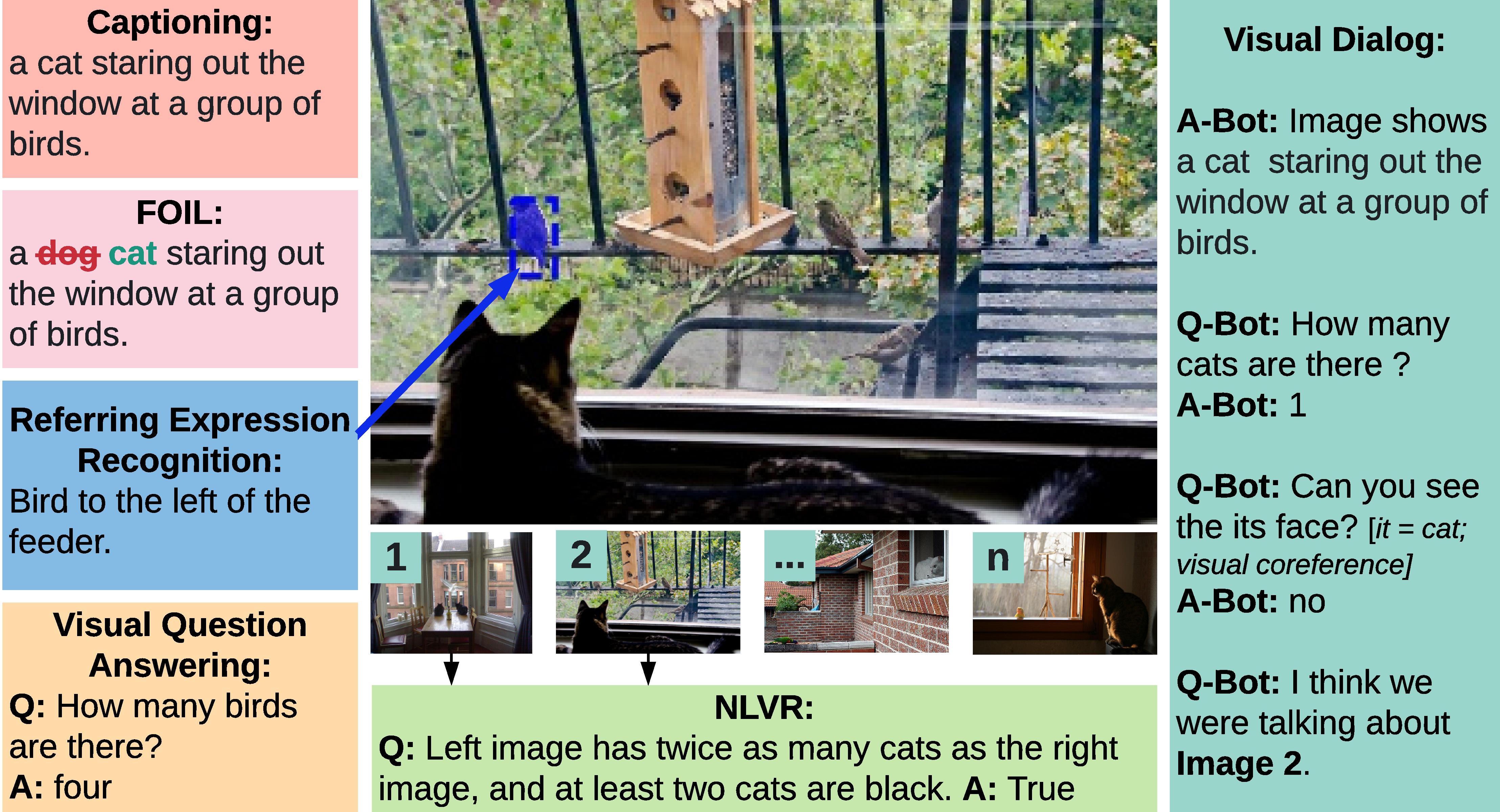}
\caption{Common tasks in vision and language research.}
\label{fig:tasks}
\end{figure}

Multiple V\&L tasks have been proposed for developing and evaluating AI systems. We briefly describe the most prominent V\&L tasks and discuss baseline and state-of-the-art algorithms. Some of these tasks are shown in Fig.~\ref{fig:tasks}.

\subsection{Tasks in V\&L research}
Bidirectional sentence-to-image and image-to-sentence retrieval problems are among the earliest V\&L tasks~\citep{mezaris2003ontology}. Early works dealt with simpler keyword-based image retrieval~\citep{mezaris2003ontology}, with later approaches using deep learning and graph-based representations~\citep{johnson2015image}. Visual semantic role labeling requires recognizing activities and semantic context in images~\citep{yatskar2016situation,Zhao2017MenAL}. Image captioning, the task of generating descriptions for visual content, involves both visual and language understanding. It requires describing the gist of the \emph{interesting} content in a scene~\citep{lin2014microsoft,donahue2015long}, while also capturing specific image regions~\citep{johnson2016densecap}. Video captioning adds the additional complexity of understanding temporal relations~\citep{yu2016video}. Unfortunately, it is difficult to evaluate the quality and relevance of generated captions without involving humans~\citep{elliott2014comparing}. Automatic evaluation metrics~\citep{papineni2002bleu,lin2004rouge,papineni2002bleu} are incapable of assigning due merit to the large range of valid and relevant descriptions for visual content and are poorly correlated with human judgment, often ranking machine-generated captions as being better than human captions~\citep{Kilickaya2017ReevaluatingAM,bernardi2016automatic}.

VQA involves answering questions about visual content. Compared to captioning, it is better suited for automatic evaluation as the output can be directly compared against ground truth answers as long as the answers are one or perhaps two words long~\citep{antol2015vqa,kumar2015ask,goyal2017making}. VQA was proposed as a form of visual Turing test, since answering arbitrary questions could demand many different skills to facilitate scene understanding. While many believed VQA would be extremely challenging, results on the first natural image datasets quickly rivaled humans, which was in large part due to question-answer distribution bias being ignored in evaluation~\citep{AgrawalBP16,Zhang_2016_CVPR,agrawal2017cvqa,agrawal2018don,kafle2017analysis}. Results were good for common questions, but systems were fragile and were incapable of handling rare questions or novel scenarios. Later datasets attempted to better assess generalization. The Task Directed Image Understanding Challenge (TDIUC) tests generalization to multiple question-types~\citep{kafle2017analysis}, Compositional VQA (C-VQA) evaluates the ability to handle novel concept compositions~\citep{agrawal2017cvqa}, and VQA under Changing Priors (VQA-CP) tests generalization to different answer distributions~\citep{agrawal2018don}. It is harder to excel on these datasets by just exploiting biases. However, the vast majority of the questions in these datasets do not require complex compositional reasoning. The CLEVR dataset attempts to address this by generating synthetic questions demanding complex chains of reasoning about synthetic scenes consisting of simple geometric shapes~\citep{johnson2017clevr}. Similar to CLEVR, the GQA dataset measures compositional reasoning in natural images by asking long and complex questions in visual scenes involving real-world complexities~\citep{hudson2019gqa}.  Video Question Answering has the additional requirement of understanding temporal dynamics~\citep{zhu2017uncovering,zhao2017video}. We refer readers to survey articles for extensive reviews on VQA~\citep{kafle2016review} and image captioning~\citep{bernardi2016automatic}.

With VQA, models do not have to provide visual evidence for their outputs. In contrast, RER requires models to provide evidence by either selecting among a list of possible image regions or generating bounding boxes that correspond to input phrases~\citep{kazemzadeh2014referitgame,rohrbach2016grounding,plummer2017phrase}. Since the output of an RER query is \textit{always} a single box, it is often quite easy to \textit{guess} the correct box. To counter this, \cite{acharya2019vqd} proposed visual query detection (VQD), a form of goal-directed object detection, where the query can have 0--15 valid boxes making the task more difficult and more applicable to real-world applications. FOIL  takes a different approach and requires a system to differentiate invalid image descriptions from valid ones~\citep{shekhar2017foil_acl}. Natural Language Visual Reasoning (NLVR) requires verifying if image descriptions are true~\citep{suhr2017corpus,suhr2018corpus}.

Unlike the aforementioned tasks, EmbodiedQA requires the agent to explore its environment to answer questions~\citep{das2018embodied}. The agent must actively perceive and reason about its visual environment to determine its next actions. In visual dialog, an algorithm must hold a conversation about an image~\citep{das2017visual,das2017visdialrl}. In contrast to VQA, visual dialog requires understanding the conversation history, which may contain visual co-references that a system must resolve correctly. The idea of conversational visual reasoning has also been explored in Co-Draw~\citep{Kim2017CoDrawVD}, a task where a \textit{teller} describes visual scenes and a \textit{drawer} draws them without looking at the original scenes.

Of course, it is impossible to create an agent that knows everything about the visual world. Agents are bound to encounter novel situations, and handling these situations requires them to be aware of their own limitations. Visual curiosity addresses this by creating agents that pose questions to knowledgeable entities, e.g., humans or databases, and then they incorporate the new information for future use~\citep{zhang2018goal,yang2018visual,misra2017lba}.

\subsection{V\&L algorithms}

There are  similarities between many V\&L algorithms. Almost all algorithms use pre-trained CNNs for natural scenes and train shallow CNNs for synthetic scenes~\citep{santoro2017simple}. For language representation, almost all models use recurrent neural networks, with LSTM~\citep{hochreiter1997long} and GRU~\citep{Cho2014OnTP} being the most popular choices. Some algorithms have also made use of linguistic parsers to discover sub-tasks in natural language queries~\citep{andreas2016learning,hu2017learning}. Recently, the community has been adopting graph-based representations for image retrieval~\citep{johnson2015image}, image generation~\citep{johnson2018image}, VQA~\citep{yi2018neural}, and  semantic knowledge incorporation~\citep{yi2018neural}, due to their intuitiveness and suitability for symbolic reasoning~\citep{johnson2015image}.

Most of the V\&L algorithms fuse visual and language representations. Fusion mechanisms range from simple techniques, such as concatenation and Hadamard products~\citep{kafle2016,antol2015vqa}, to more intricate methods, e.g., bilinear fusion~\citep{FukuiPYRDR16}, which are argued to better capture interactions between visual and linguistic representations. Attention mechanisms that enable extraction of query-relevant information have also been heavily explored~\citep{Yang2016,Anderson2017up-down,Kim2018BilinearAN,yu2018mattnet}. Attention mechanisms learn to assign higher \textit{importance} to relevant information using both top-down and bottom-up pathways~\citep{Anderson2017up-down}.

Another common requirement for many V\&L tasks is a multi-step reasoning mechanism. For this, the community has proposed modular networks that use pre-defined components to perform pre-specified reasoning functions, e.g., filtering and describing visual regions~\citep{andreas2016learning,hu2017learning,yu2018mattnet}, providing a transparent reasoning process. Compositional reasoning can also be achieved by capturing pairwise interactions between V\&L representations~\citep{santoro2017simple} and by recurrently extracting and consolidating information from the input~\citep{arad2018compositional}. These approaches directly learn reasoning from data by utilizing structural biases provided by the model definition.

While these algorithms show impressive new capabilities, their development and evaluation has been split into two distinct camps: the first camp focuses on monolithic architectures that often excel at natural image V\&L tasks~\citep{Yang2016,kim2016multimodal},  whereas the second camp focuses on compositional architectures, that excel at synthetically generated scenes testing for compositional reasoning~\citep{santoro2017simple,arad2018compositional}. Algorithms developed for one camp are often not evaluated on the datasets from other camp, which makes it difficult to gauge the true capabilities of V\&L algorithms. \cite{shrestha2019ramen} showed that most of the algorithms developed for natural image VQA do not perform well on synthetic compositional datasets and vice-versa. The authors further propose a simple architecture that compares favorably against state-of-the-art algorithms from both camps, indicating that specialized mechanisms such as: attention, modular reasoning and fusion mechanisms, used in more intricate methods may been been over-engineered to perform well on selected datasets.

\section{Shortcomings of V\&L research}\label{sec:shortcomings}

Progress in V\&L has been swift. Benchmarks for several V\&L tasks show that algorithms are equalling or even surpassing human performance~\citep{Johnson2017InferringAE,bernardi2016automatic}. In this section, we will outline several shortcomings and challenges faced by the V\&L tasks that show that existing results can be misleading. We will also discuss the efficacy of existing remedial measures in tackling these shortcomings.

\subsection{Dataset bias}\label{sec:shortcomings-bias}

  
		

\begin{figure}[t]
		\includegraphics[width=\textwidth]{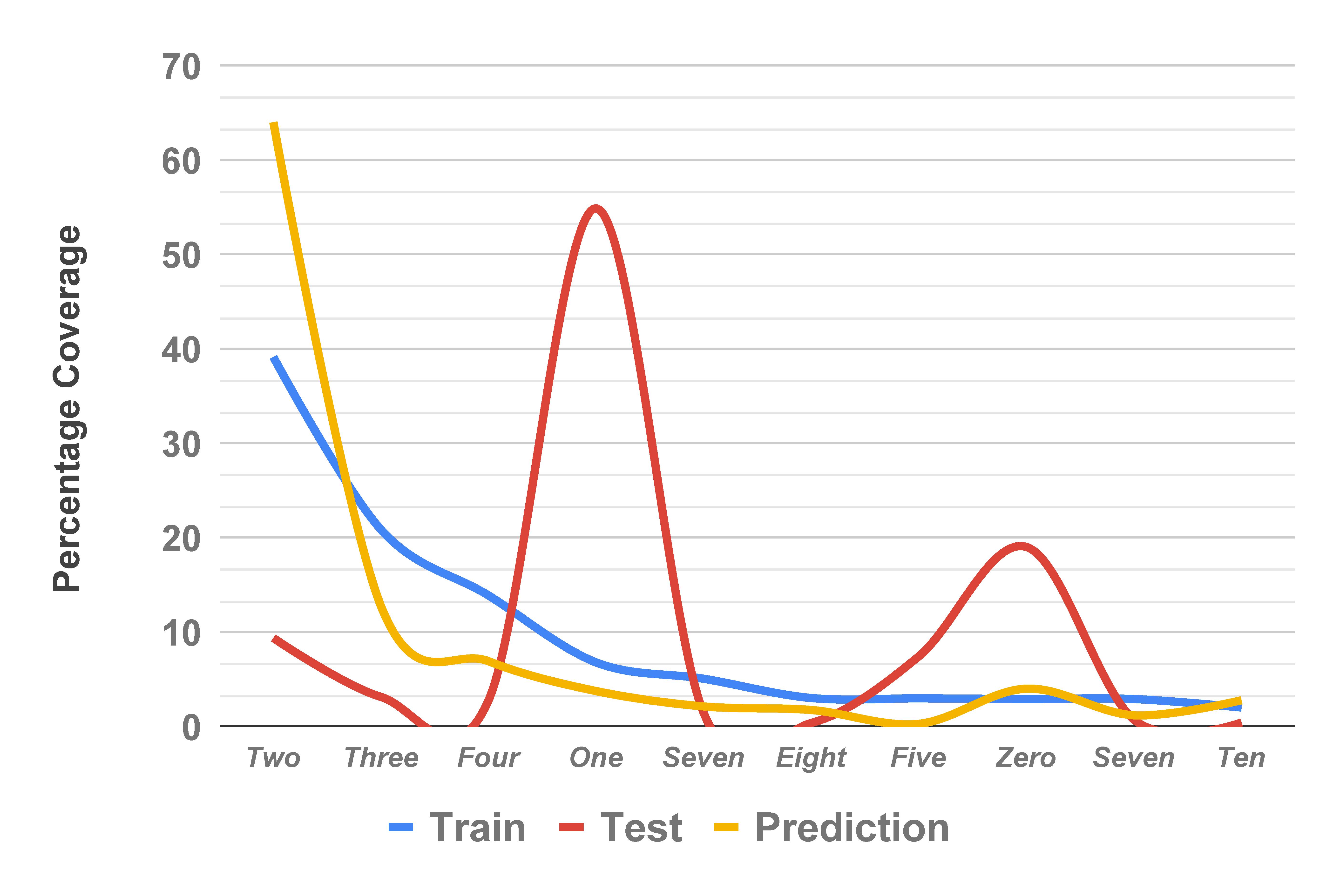}
        \caption{Distribution of answers for questions starting with `How many' in the train and test split of the VQA-CP dataset~\citep{agrawal2018don}, and test-set predictions of the state-of-the-art VQA model, BAN~\citep{Kim2018BilinearAN}. In VQA-CP, the distribution of test is intentionally made different from train to assess if the algorithms can perform under changing priors. Algorithms not only fails to perform well under changing priors, but it also has a bias-amplification effect, i.e., the predictions show even stronger bias towards answers that are more common in the training set than the actual level of bias. Similar observations have been made for semantic role labeling~\citep{Zhao2017MenAL}. \label{fig:bias-amp}}
\end{figure}

Unwanted or unchecked biases in natural datasets are arguably the most prevalent issues in V\&L tasks. Since the data used for training and testing a model are often collected homogeneously~\citep{antol2015vqa,goyal2017making,lin2014microsoft}, they share common patterns and regularities. Hence, it is possible for an algorithm to get good results by memorizing those patterns, undermining our efforts to evaluate the understanding of vision and language. The biases in datasets can stem from several sources, can be hard to track, and can result in severely misleading model evaluation. Two of the most common forms of bias stem from bias in crowd-sourced annotators and naturally occurring regularities. Finally, `photographer's bias' is also prevalent in V\&L benchmarks, because images found on the web share similarities in posture and composition due to humans having preferences for specific views~\citep{azulay2018deep}. Since the same biases and patterns are also mirrored in the test dataset, algorithms can simply memorize these superficial patterns (If the question has the pattern `Is there an \texttt{OBJECT} in the picture?', then answer `yes') instead of learning to actually solve the intended task (answer `yes' only if the \texttt{OBJECT} is actually present). If this bias is not compensated for during evaluation, benchmarks may only test a very narrow subset of capabilities. This can enable algorithms to perform well for the wrong reasons and algorithms can end up catastrophically failing in uncommon scenarios~\citep{alcorn2018strike,agrawal2018don}.

Several studies demonstrate the issue of bias in V\&L tasks. For example, blind VQA models that `guess' the answers without looking at images show relatively high accuracy~\citep{kafle2016}. In captioning, simple nearest neighbor-based approaches yield surprisingly good results~\citep{devlin2015exploring}. Dataset bias occurs in other V\&L tasks as well~\citep{Zhao2017MenAL,shekhar2017foil_acl,zellers2018neural,cirik2018visual}. Recent studies~\citep{Zhao2017MenAL} have shown that algorithms not only \textit{mirror} the dataset bias in their predictions, but in fact \textit{amplify} the effects of bias (see Fig.~\ref{fig:bias-amp}).

Numerous studies have sought to quantify and mitigate the effects of answer distribution bias on an algorithm's performance. As a straightforward solution, \cite{Zhang_2016_CVPR} and \cite{kafle2017analysis}  proposed balanced training sets with a uniform distribution over possible answers. This is somewhat effective for simple binary questions and synthetically generated visual scenes, but it does not address the imbalance in the kinds of questions present in the datasets. Re-balancing all kinds of query types is infeasible for large-scale natural image datasets. Furthermore, it may be counterproductive to forgo information contained in natural distributions in the visual and linguistic content, and focus should instead be on rigorous evaluation that compensates for bias or demonstrates bias robustness~\citep{agrawal2018don}. We discuss this further in the next section.

\subsection{Evaluation metrics}\label{sec:shortcomings-evaluation}

Proper evaluation of V\&L algorithms is difficult. A lot of the challenge arises from the complexity of the natural language. Language can be used to express similar semantic content in different ways, which makes automatic evaluation of models that emit words and sentences particularly challenging. For example, the captions `A man is walking next to a tree' and `A guy is taking a stroll by the tree' are nearly identical in meaning, but it can be hard for automatic systems to infer that fact. Several evaluation metrics have been proposed for captioning, including simple n-gram matching systems (e.g., BLEU~\citep{papineni2002bleu}, CIDEr~\citep{vedantam2015cider} and ROUGE~\citep{lin2004rouge}) and human consensus-based measures~\citep{vedantam2015cider}. Most of these metrics have limitations~\citep{Kilickaya2017ReevaluatingAM,bernardi2016automatic}, with n-gram based metrics suffering immensely for sentences that are phrased differently but have identical meaning or use synonyms~\citep{Kilickaya2017ReevaluatingAM}. Alarmingly, evaluation metrics often rank machine-generated captions as being better than human captions but fail when human subjectivity is taken into account~\citep{bernardi2016automatic,Kilickaya2017ReevaluatingAM}. Even humans find it hard to agree on what a `good' caption entails~\citep{vedantam2015cider}. Automatic evaluation of captioning is further complicated because it is not clear what is expected from the captioning system. A given image can have many valid captions ranging from descriptions of specific objects in an image, to an overall description of the entire image. However, due to natural regularities and photographer bias, generic captions can apply to a large number of images, thereby gaining high evaluation scores without demonstrating visual understanding~\citep{devlin2015exploring}.

Evaluation issues are lessened in VQA and RER where the output is better defined; however, it is not completely resolved. If performance for VQA is measured using exact answer matches, then even small variations will be harshly punished, e.g., if a model predicts `bird' instead of `eagle', then the algorithm is punished as harshly as if it were to predict `table.' Several solutions have been proposed, but they have their own limitations, e.g., Wu-Palmer Similarity (WUPS), a word similarity metric, cannot be used with sentences and phrases. Alternately, consensus based metrics have been explored ~\citep{malinowskiRF15,antol2015vqa}, where multiple annotations are collected for each input, with the intention of capturing common variations of the ground truth answer. However, this paradigm can make many questions \textit{unanswerable} due to low human consensus~\citep{kafle2017analysis,kafle2016}. Multiple-choice evaluation has been proposed by several benchmarks~\citep{antol2015vqa,goyal2017making}. While this simplifies evaluation, it takes away a lot of the open-world difficulty from the task and can lead to inflated performance via smart guessing~\citep{jabri2016revisiting}.

Dataset biases introduce further complications for evaluation metrics. Inadequate metrics can conflate the issues of bias when the statistical distributions of the training and test sets are not taken into account, artificially inflating performance. Metrics normalized to account for the distribution of training data \citep{kafle2017analysis} and diagnostic datasets that artificially perturb the distribution of train and test data~\citep{agrawal2018don} have been proposed to remedy this. Furthermore, open-ended V\&L language tasks can \emph{potentially} test a variety of skills, ranging from relatively easy sub-tasks (detection of large, well-defined objects), to fairly difficult sub-tasks (fine-grained attribute detection, spatial and compositional reasoning, counting, etc.). However, these tasks are not evenly distributed. Placing all skill types on the same footing can inflate system scores and hide how fragile these systems are. Dividing the dataset into underlying tasks can help~\citep{kafle2017analysis}, but the best way to make such a division is not clearly defined.

\subsection{Are V\&L systems `horses?'}\label{sec:shortcomings-horses}

\begin{figure}[t]
\centering
\includegraphics[width=\textwidth]{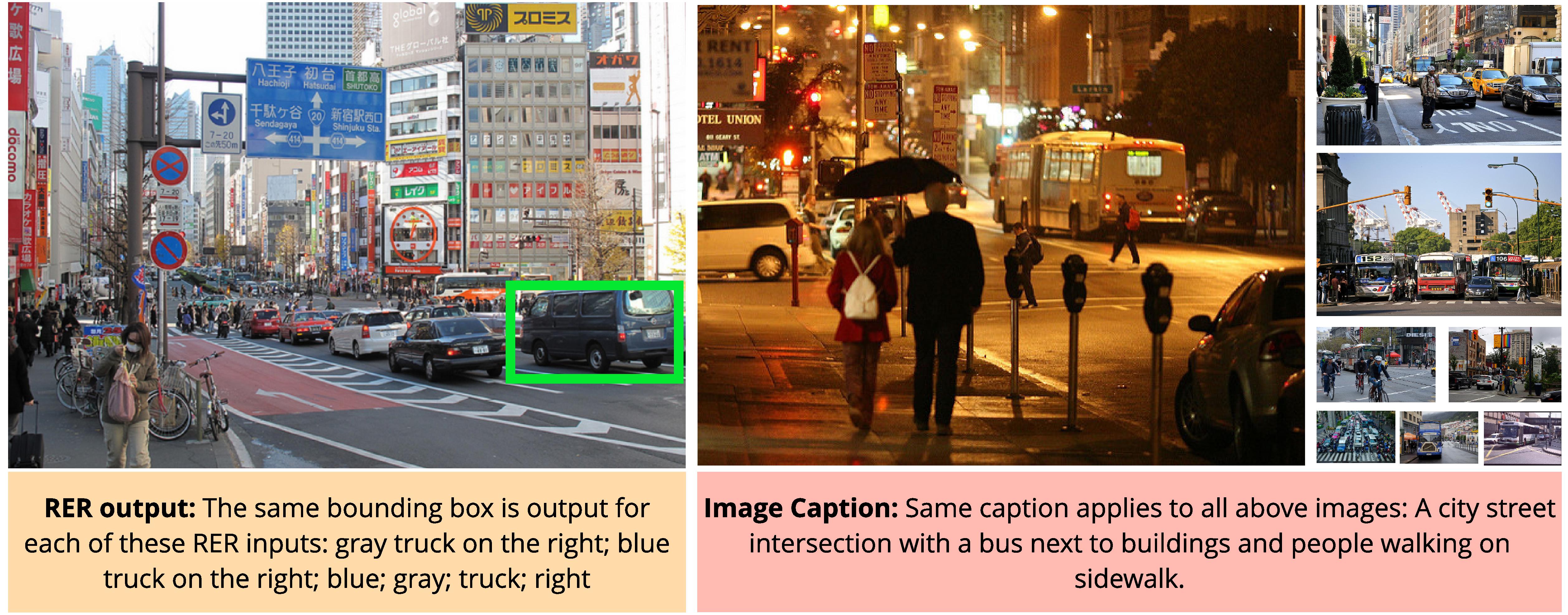}
\caption{The apparent versus true complexity of V\&L tasks. In RER (left), omitting a large amount of text has no effect on the output of the system~\citep{yu2018mattnet}.  Similarly, a seemingly detailed caption (right) can apply to a large number of images from the dataset making it easy to `guess' based on shallow correlations. While it appears as though the captioning system can identify objects (`bus', `building', `people'), spatial relationships (`next to', `on'), and activities (`walking'). However, it is entirely possible for the captioning system to have `guessed' the caption by detection of one of the objects in the caption, e.g., a `bus' or even \textit{a common latent} object such as `traffic light'.}
\label{fig:shortcomings}
\end{figure}

Strum defines a `horse' as \textbf{`a system that appears as if it is solving a particular problem when it actually is not'}~\citep{sturm2016horse}. Of course, the `horse' here refers to the infamous horse named Clever Hans, thought to be capable of arithmetic and abstract thought but was in reality exploiting the micro-signals provided by its handler and audience. Apart from bias and evaluation, there are other issues in V\&L datasets that are harder to pinpoint. We review several of these issues and highlight existing studies that scrutinize the true capabilities of existing V\&L systems to assess whether they are `horses.'

\subsubsection{Superficial correlations and true vs. apparent difficulty}

Due to superficial correlations, the difficulty of V\&L datasets may be much lower than the true difficulty of comprehensively solving the task (see Fig.~\ref{fig:shortcomings}). We outline some of the key studies and their findings that suggest  V\&L algorithms are relying on superficial correlations that enable them to achieve high performance in common situations but make them vulnerable when tested under different, but not especially unusual, conditions.

\textbf{VQA:} Image-blind algorithms that only see questions often perform surprisingly well~\citep{kafle2016,Yang2016}, sometimes even surpassing the algorithms having access to both ~\citep{kafle2016}. Algorithms also often provide inconsistent answers due to irrelevant changes in phrasing~\citep{kafle2016review,ray2018make}, signifying  a lack of question comprehension. When a VQA dataset is divided into different question-types, algorithms performed well only on easier tasks that CNNs alone excel at, e.g., detecting whether an object is present, but they performed  poorly for complex questions that require bi-modal reasoning~\citep{kafle2017analysis}. This discrepancy in accuracy is not clearly conveyed when simpler accuracy metrics are used. In a multi-faceted study, \cite{AgrawalBP16} showed several quirks of VQA, including how VQA algorithms converge to an answer without even processing one half of the question and show an inclination to fixate on the same answer when the same question is repeated for a different image. Similarly, \cite{goyal2017making} showed that VQA algorithm performance deteriorates when tested on pairs of images that have opposite answers. As shown in Fig.~\ref{fig:bias-amp}, VQA systems can actually amplify bias.

\textbf{Image captioning:} In image captioning, simply predicting the caption of the training image with the most similar visual features yields relatively high scores using automatic evaluation metrics~\citep{devlin2015exploring}. Captioning algorithms exploit multi-modal distributional similarity~\citep{madhyastha2018end}, and generate captions similar to images in the training set, rather than learning concrete representations of objects and their properties.

\textbf{Embodied QA and visual dialog:} EmbodiedQA ostensibly requires navigation, visual information collection, and reasoning, but \cite{anand2018blindfold} showed that vision blind algorithms perform competitively. Similarly, visual dialog \textit{should} require understanding both visual content and dialog history~\citep{massiceti2018visual}, but an extremely simple method produces near state-of-the-art performance for visual dialog, despite ignoring both visual and dialog information~\citep{massiceti2018visual}.

\textbf{Scene graph parsing:} Predicting scene graphs requires understanding object properties and their relationships to each other. However, \cite{zellers2018neural} showed that objects alone are highly indicative of their relationship labels. They further demonstrated that for a given object pair, simply guessing the most common relation for those objects in the training set yields improved results compared to state-of-the-art methods.

\textbf{RER:} In a multi-faceted study of RER, \cite{cirik2018visual} demonstrated multiple alarming issues. The first set of experiments involved tampering with the input referring expression to examine if algorithms properly used the text information. Tampering should reduce performance if algorithms make proper use of text to predict the correct answers. However, their results were relatively unaffected when the words were shuffled and nouns/adjectives were removed from the referring expressions. This signifies that it is possible for algorithms to get high scores without explicitly learning to model the objects, attributes and their relationships. The second set of experiments demonstrated that it is possible to predict correct candidate boxes for over 86\% of referring expressions, without ever feeding the referring expression to the system. This demonstrates that algorithms can exploit regularities and biases in these datasets to achieve good performance, making these datasets a poor test of the RER task. 

Some recent works have attempted to create more challenging datasets that probe the abilities to properly ground vision and language beyond shallow correlations. In FOIL~\citep{shekhar2017foil_acl}, a single noun from a caption is replaced with another, making the caption invalid. Here the algorithm, must determine if the caption has been \textit{FOILed} and then detect the \textit{FOIL} word and replace it with a correct word. Similarly, in NLVR~\citep{suhr2017corpus}, an algorithm is tasked with finding whether a description applies to a pair of images. Both tasks are extremely difficult for modern V\&L algorithms with the best performing system on NLVR limited to around 55\% (random guess is 50\%), well short of the human performance of over 95\%. These benchmarks may provide a challenging test bed that can spur the development of next-generation V\&L algorithms. However, they remain limited in scope, with FOIL being restricted to noun replacement for a small number of categories (less than 100 categories from the COCO dataset). Hence, it does not test understanding of attributes or relationships between objects. Similarly, NLVR is difficult, but it lacks additional annotations to aid in the measurement of \textit{why} a model fails, or eventually, why it succeeds.

\subsubsection{Lack of interpretability and confidence}\label{sec:shortcomings-interpretability}

Human beings can provide explanations, point to evidence, and convey confidence in their predictions. They also have an ability to say `I do not know' when the information provided is insufficient. However, almost none of the existing V\&L algorithms are equipped with these abilities, making the models highly uninterpretable and unreliable. 

In VQA, algorithms provide high-confidence answers even when the question is nonsensical for a given image, e.g., `What color is the horse?' for an image that does not contain a horse can yield `brown' with a very high confidence. Very limited work has been done in V\&L to assess a system's ability to deal with lack of information. While \cite{kafle2017analysis} proposed a class of questions called `absurd' questions to test a system's ability to determine if a question was unanswerable, they were limited in scope to simple detection questions. More complex forms of absurdity are yet to be tested. 

Because VQA and captioning do not explicitly require or test for proper grounding or pointing to evidence, the predictions made by these algorithms remain uninterpretable. A commonly practiced remedy is to include visualization of attention maps for attention-based methods, or use post-prediction visualization methods such as Grad-CAM~\citep{selvaraju2017grad}. However, these visualizations shed little light on whether the models have `attended' to the right image regions. First, most V\&L datasets do not contain attention maps that can be compared to the predicted attention maps; therefore, it is difficult to gauge the prediction quality. Second, even if such data were available, it is not clear what image regions the model \textit{should} be looking at. Even for well-defined tasks such as VQA, answers to questions like `Is it sunny?' can be inferred using multiple image regions. Indeed, inclusion of attention maps does not make a model more predictable for human observers~\citep{chandrasekaran2018explanations}, and the attention-based models and humans do not \textit{look} at same image regions~\citep{das2016human}. This suggests attention maps are an unreliable means of conveying interpretable predictions.

Several works propose the use of textual explanations to improve interpretability \citep{hendricks2016generating,li2018vqa}. \cite{li2018vqa} collected text explanations in conjunction with standard VQA pairs and a model must predict both the correct answer and the explanation. However, learning to predict explanations can suffer from many of the same problems faced by image captioning: evaluation is difficult and there can be multiple  valid explanations. Currently, there is no reliable evidence that such explanations actually make the model more interpretable, but there is some evidence of the contrary~\citep{chandrasekaran2018explanations}.

Modular and compositional approaches attempt to reveal greater insight by incorporating interpretability directly into the design of the network~\citep{hu2017learning,johnson2018image,Johnson2017InferringAE}. However, these algorithms are primarily tested on simpler, synthetically constructed datasets that lack the diversity of natural images and language. The exceptions that are tested on natural images rely on hand-crafted semantic parsers to pre-process the questions~\citep{hu2017learning}, which often over-simplify the complexity of the questions~\citep{kafle2016review}.

\subsubsection{Lack of compositional concept learning}\label{sec:shortcomings-compositionality}

It is hard to verify that a model has understood concepts. One method to do this is to use it in a novel setting or in a previously unseen combination. For example, most humans would not have a problem recognizing a purple colored dog, even if they have never seen one before, given that they are familiar with the concepts of purple and dog. Measuring such compositional reasoning could be crucial in determining whether a  V\&L system is a `horse.' This idea has received little attention, with few works devoted to it~\citep{johnson2017clevr,agrawal2017cvqa}. Ideally, an algorithm should not show any decline in performance for novel concept combinations. However, even for CLEVR, which is composed of basic geometric shapes and colors, most algorithms show a large drop in performance for novel shape-color combinations~\citep{johnson2017clevr}. For natural images, the drop in performance is even higher~\citep{agrawal2017cvqa}.

\section{Addressing Shortcomings}\label{sec:prospects}

\begin{table}[t]
\centering
\caption{A summary of challenges and potential solutions for V\&L problems.}
\label{table:summary}
\begin{tabular}{ll@{}}
\toprule
 \multicolumn{1}{c}{\textbf{Shortcomings/Challenges}}                                                                                                                                 & \multicolumn{1}{c}{\textbf{Potential Solutions}}                                                                                                                                                                              \\ \midrule

 \begin{tabular}[c]{p{6cm}} 
Evaluation metrics are a poor measure for competence of algorithms due to dataset bias.
 \end{tabular}                                 & \begin{tabular}[c]{p{8cm}}\tabitem Use metrics that account for dataset biases.\\ \tabitem Carefully measure and report performance on individual abilities.
\end{tabular} \\ \midrule

\begin{tabular}[c]{p{6cm}} 
It is hard to tell if algorithms are `right for the right reasons.' They can perform well on benchmarks without actually solving the problem.
 \end{tabular}                                 & \begin{tabular}[c]{p{8cm}}\tabitem Test the algorithms by withholding varying degrees of task-critical information from them to measure if they understand concepts. \\ \tabitem Measure task understanding by asking the model to do the same task in dissimilar contexts and with alternative phrasing. \\ \tabitem Develop defense mechanisms against `accidentally' reaching the correct solutions.
\end{tabular} \\ \midrule

\begin{tabular}[c]{p{6cm}} 
Trained systems are fragile and easily break when humans use them.
 \end{tabular}                                 & \begin{tabular}[c]{p{8cm}}\tabitem Incorporate prediction confidence into evaluation. 
\\ \tabitem Allow systems to output `I don’t know.'
\end{tabular} \\ \midrule

\begin{tabular}[c]{p{6cm}} 
V\&L Systems are one-trick-ponies, rarely able to generalize to more than one task.
 \end{tabular}                          & \begin{tabular}[c]{p{8cm}}\tabitem Create a V\&L decathlon that tests  numerous V\&L tasks. Assess positive transfer among tasks.
  \end{tabular} \\

\\ \bottomrule
\end{tabular}
\end{table}

In this survey, we have compiled a wide range of shortcomings and challenges faced by modern V\&L research based on the datasets and evaluation of tasks. One of the major issues stems from the difficulty in evaluating if an algorithm is actually solving the task, which is confounded by hidden perverse incentives in modern datasets that cause algorithms to exploit unwanted correlations. Lamentably, most proposed tasks do not have built-in safeguards against this or even an ability to measure it. Many \textit{post-hoc} studies have shed light on this problem. However, they are often limited in scope, require collecting additional data~\citep{shekhar2017foil_acl}, or the modification of `standard' datasets~\citep{agrawal2017cvqa,agrawal2018don,kafle2016}. We outline prospects for future research in V\&L, with an emphasis on discussing the characteristics of future V\&L tasks and evaluation suites that are better aligned with the goals of a visual Turing test. Table~\ref{table:summary} presents a short summary of challenges and potential solutions in V\&L research.

\subsection{New natural image tasks that measure core abilities}

Existing V\&L evaluation schemes for natural datasets ignore bias, making it possible for algorithms to excel on standard benchmarks without demonstrating proper understanding.  We argue that a carefully designed suite of tasks could be used to overcome this obstacle.  We propose some possible approaches to improve evaluation by tightly controlling the evaluation of core abilities and ensuring that evaluation compensates for bias.

\begin{figure}[t]
\centering  \includegraphics[width=0.6\textwidth]{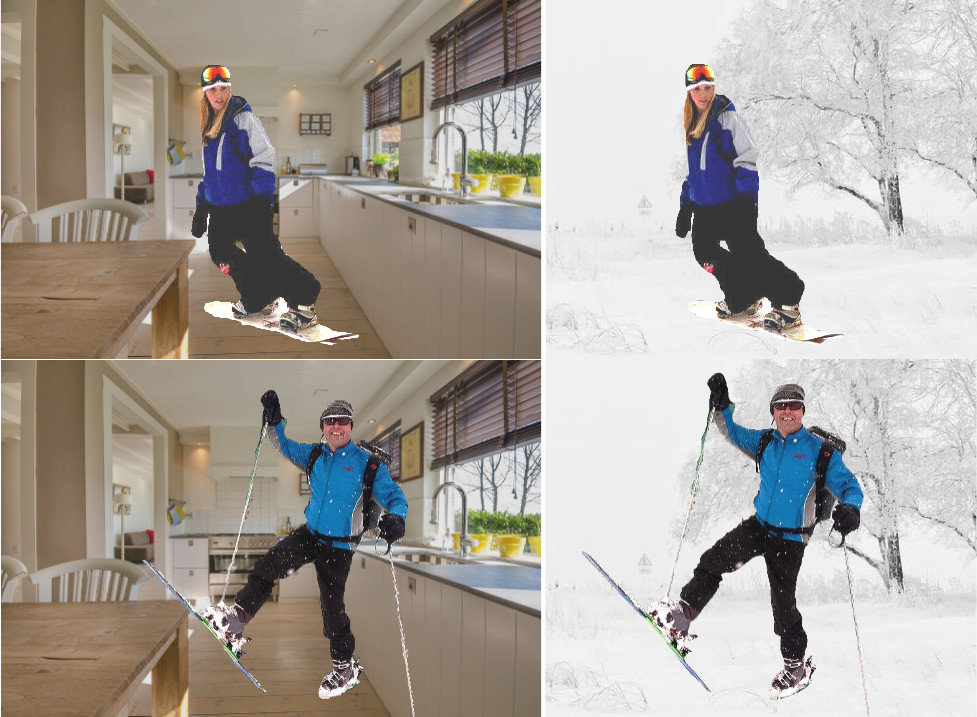}
    \caption{\emph{Posters} dataset can help test bias.
  In this example, both contextual and gender bias are tested by placing  out-of-context poster-cut-outs. Snowboarding is generally correlated with gender `male' and context `snow'~\citep{Hendricks2018WomenAS}.
  }
  \label{fig:posters}
\end{figure}

Programmatically created datasets, e.g., CLEVR for VQA, can enable fine-grained evaluation of specific components by using simple synthetically created scenes. We could create a similar dataset for natural images by composing scenes of natural objects (see Fig.~\ref{fig:posters}). This could be used to test higher-levels of visual knowledge, which is not possible in synthetic environments. This approach could be used to examine reasoning and bias-resistance by placing objects in unknown combinations and then asking questions with long reasoning chains, novel concept compositions, and distinct train/test distributions.

Current benchmarks cannot reliably ascertain whether an algorithm has learned to represent objects and their attributes properly, and it is often easy to produce a correct response by `guessing' prominent objects in the scene~\citep{cirik2018visual}. To examine whether an algorithm demonstrates concept understanding, we envision a dataset containing simple queries, where given a set of objects and/or attributes as queries, the algorithm needs to highlight \textit{all} objects that satisfy \textit{all} of the conditions in the set, e.g., for \textit{query=\{{red}\}}, the algorithm must detect all red objects, and for \textit{\{{red,car}\}}, it must detect all red cars. However, all queries would have \textit{distractors} in the scene, e.g., \{\textit{red, car}\} is only used when the scene also contains 1) cars that are non-red, 2) objects other than cars or 3) other non-red objects. By abandoning the complexity of natural language, this dataset allows for the creation of queries that are hard to `guess' without learning proper object and attribute representations. Since the chance of a random guess being successful is inversely proportional to number of \textit{distractors}, the scoring can also be made proportional to \textit{additional} information over a random guess.

We hope that carefully designed test suites that measure core abilities of V\&L systems in a controlled manner will be developed. This serves as a necessary adjunct to more open-ended benchmarks, and would help dispel the `horse' in V\&L.

\subsection{Better evaluation of V\&L systems}\label{sec:prospect-evaluation}

V\&L needs better evaluation metrics for standard benchmarks. Here, we will outline some of the key points future evaluation metrics should account for:

\begin{itemize}
  \setlength\itemsep{0em}
    \item Evaluation should test individual skills to account for dataset biases~\citep{kafle2017analysis} and measure performance relative to  `shallow' guessing~\citep{agrawal2018don,kafle2016review,cirik2018visual}.
    \item Evaluation should include built-in tests for `bad' or `absurd' queries~\citep{kafle2017analysis,cirik2018visual}.
    \item Test sets should contain a large number of compositionally novel instances that can be inferred from training but not directly matched to a training instance~\citep{devlin2015exploring,johnson2017clevr}.
    \item Evaluation should keep the `triviality' of the task in mind when assigning score to a task, e.g., if there is only a single cat then `Is there a black cat sitting between the sofa and the table?' reduces to `Is there a cat?' for that image~\citep{AgrawalBP16,cirik2018visual}.
    \item Robustness to semantically identical queries must be assessed.
    \item Evaluation should be done on questions with unambiguous answers; if humans strongly disagree, it is likely not a good question for a visual Turing test.
\end{itemize}
We believe future evaluation should probe algorithms from multiple angles such that a single score is derived from a suite of sub-scores that test different capabilities. The score could be divided into underlying core abilities (e.g., counting, object detection, fine-grained recognition, etc.), and also higher-level functions (e.g., consistency, predictability, compositionality, resistance to bias, etc.)

\subsection{V\&L decathlon}

Most of the V\&L tasks seek to measure language grounded visual understanding. Therefore, it is not unreasonable to expect an algorithm designed for one benchmark to readily transfer to other V\&L tasks with only minor modifications. However, most algorithms are tested on single task~\citep{kafle2016,yu2018mattnet,Yang2016}, with very few exceptions~\citep{Anderson2017up-down,Kim2018BilinearAN,shrestha2019ramen}. Even within the same task, algorithms are almost never evaluated on multiple datasets to assess different skills, which makes it difficult to study the true capabilities of the algorithms. 

To measure holistic progress in V\&L research, we believe it is imperative to create a large-scale V\&L decathlon benchmark. Work in a similar spirit has recently been proposed as DecaNLP~\citep{McCann2018decaNLP}, where many constituent NLP tasks are represented in a single benchmark. In DecaNLP, all constituent tasks are represented as question-answering for an easier input-output mapping. To be effective, a V\&L decathlon benchmark should not only contain different sub-tasks and diagnostic information but also entirely different input-output paradigms. We envision models developed for a V\&L decathlon to have a central V\&L core and multiple input-output nodes that the model selects based on the input. Both training and test splits of the decathlon should consist of many different input-output mappings representing distinct V\&L tasks. For example, the same image could have a \textbf{VQA question} `What color is the cat?', a \textbf{pointing question} `What is the color of ``that'' object?', where ``that'' is a bounding box pointing to an object, and a \textbf{RER} `Show me the red cat.' Integration of different tasks encourages development of more capable V\&L models. Finally, the test set should contain unanswerable queries~\citep{kafle2017analysis,cirik2018visual}, compositionally novel instances~\citep{Johnson2017InferringAE,agrawal2017cvqa}, pairs of instances with subtle differences~\citep{goyal2017making}, equivalent queries with same ground truth but different phrasings, and many other quirks that allow us to peer deeper into the reliability and true capacity of the models. These instances can then be used to produce a suite of metrics as discussed earlier. 

\section{Conclusion}

While V\&L work originally seemed incredibly difficult, progress on benchmarks rapidly made it appear that systems would soon rival humans. A wide range of studies have shown that much of this progress may be misleading due to flaws in standard evaluation methods. While this should serve as a cautionary story for future research in other areas, V\&L research has a bright future. While the vast majority of research is on creating new algorithms, we argue that constructing good evaluation techniques is just as critical, if not more so, for progress to continue. V\&L has the potential to be a visual Turing test for assessing progress in AI, but this potential can only be achieved through the monumental task of developing strong benchmarks that evaluate many capabilities individually and thoroughly on rich real-world imagery to evaluate system competence. These systems can enable much richer interactions with computers and robots, but this demands that the systems are trustworthy and robust across scenarios.

\section*{Conflict of Interest Statement}

Author CK was employed by commercial company PAIGE. All other authors declare no competing interests. 


\section*{Author Contributions}

CK and KK conceived of the scope of this review. KK, RS, and CK all contributed to the text of this manuscript.


\section*{Funding}
This work was supported in part by a gift from Adobe Research.

\section*{Acknowledgments}
We thank Manoj Acharya, Ryne Roady, and Tyler Hayes for comments and review of this paper.


\bibliographystyle{frontiersinSCNS_ENG_HUMS} 
\bibliography{frontiers}



\end{document}